\documentclass[letterpaper, 10 pt, conference]{ieeeconf}  

\IEEEoverridecommandlockouts                              

\usepackage{graphics} 
\usepackage{epsfig} 
\usepackage{mathptmx} 
\usepackage{times} 
\usepackage{amsmath} 
\usepackage{amssymb}  
\usepackage[font=footnotesize]{caption,subcaption}
\usepackage{tikz}
\usepackage{amsmath}
\usepackage{url}
\DeclareMathOperator*{\argmin}{arg\,min}

\title{\LARGE \bf
Towards Autonomous Industrial-Scale Bathymetric Surveying}

\author{Ignacio Torroba\and Nils Bore\and John Folkesson
\thanks{The authors are with the Division of Robotics, Perception and Learning at KTH Royal Institute of Technology, SE-100 44 Stockholm, Sweden
        {\tt\small \{torroba, nbore, johnf\}@kth.se}}%
}  

\begin{document}

\maketitle
\thispagestyle{empty}
\pagestyle{empty}

\begin{abstract}
Both higher efficiency and cost reduction can be gained from automating bathymetric surveying for offshore applications such as pipeline, telecommunication or power cables installation and inspection on the seabed.  
We present a SLAM system that optimizes the geo-referencing of bathymetry surveys by fusing the dead-reckoning sensor data from the surveying vehicle with constraints from the maximization of the geometric consistency of overlapping regions of the survey.  
The framework has been extensively tested on bathymetric maps from both simulation and several actual industrial surveys and has proved robustness over different types of terrain.  
We demonstrate that our system is able to maximize the consistency of the final map even when there are large sections of the survey with reduced topographic variation.
The framework has been made publicly available together with the simulation environment used to test it and some of the datasets. 
\end{abstract}

\section{INTRODUCTION}

Bathymetric surveys are an essential service on a great variety of offshore industrial operations.
Tasks such as laying pipelines, setting transoceanic cables or installing wind farms require a detailed map of the topology of the seabed for a safe installation.  Furthermore, the high security standards applied on these industries make regular inspections of the seabed structures necessary for maintenance. Sound navigation ranging (sonar) is the primary tool for underwater sensing, with the multibeam echo sounder (MBES) being commonly used due to its resolution, range and robustness.

Nowadays, infrastructure assessment operations are carried out by specialized vessels that monitor large areas of seabed through MBES.
These vessels employ hull-mounted sonars in shallow waters and remotely operated vehicles (ROV) for deeper areas.
The resolution of hull-mounted equipment is limited by the depth of the target area while the ROV tethers complicate the operations.  Furthermore, geo-referencing is more challenging when the sonar is mounted on an ROV as there is no rigid connection to a GPS antenna.  

Surveys are time consuming and costly. 
Reducing the operation time offshore or the crew involved could increase the overall efficiency.  This motivates the use of autonomous underwater
vehicles (AUV) as a means to automate bathymetric surveying tasks.  
However, the long range and often large depths at which these operations
take place often pose a challenge to an AUV's navigation system.  In
the absence of fixed reference points, the error in  position of
the vehicle will grow boundlessly with the length of the mission, limiting the accuracy of the constructed map.  Post processing can correct some errors but optimization of the vehicle trajectory estimate is important to ensure full area coverage, as well as creating maps that meet the accuracy requirements.

\begin{figure}[t]
	\centering
	\includegraphics[width=0.5\textwidth]{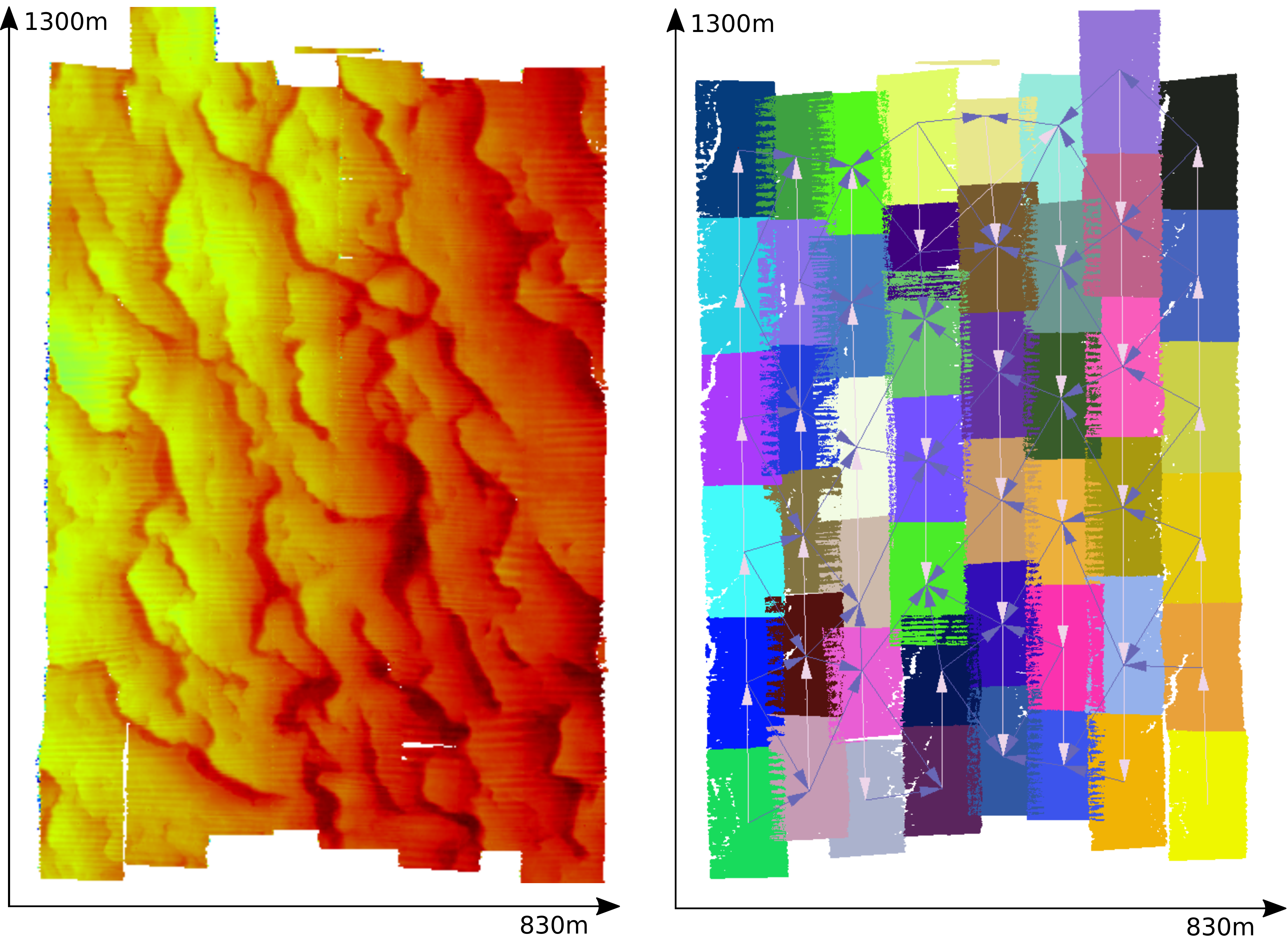}
	\caption{The depth map of the Ripples dataset and resulting pose graph with bathymetric submaps.  The trajectory of the vehicle provided constraints between consecutive maps as shown by the white arrows.  Overlapping regions provided additional constraints, shown with blue edges. }
	\label{fig:ripples_map}
\end{figure}

Installation of underwater acoustic positioning systems for terrain-aided navigation such as long baseline (LBL), when an option, would increase the complexity and overall costs.   On the other hand, systems like ultra short baseline (USBL), while easy to install, require a support vessel and fail to provide an accurate positioning at high depths.  In fact, all acoustic positioning systems suffer from inaccuracies at longer distances due to propagation effects.    

Proposed solutions to these challenges have come in the form of simultaneous localization and mapping (SLAM) algorithms for the AUVs \cite{leonard2016autonomous}.
SLAM solutions have shown robust and reliable vehicle trajectory estimation and map construction on a variety of underwater scenarios.
Nevertheless, most of these techniques rely up to a certain degree on distinguishable surroundings and carefully designed trajectories to be able to perform data association and loop closure detection. 
Hence, few attempts have been made to develop and test SLAM solutions for navigating unstructured and featureless sea bottom areas on long range missions.

We present a SLAM framework designed to deal with the precision standards of industrial bathymetric surveying with underwater vehicles.  Our system, described in Sec. \ref{sec:slam_bathymetry}, integrates components from established SLAM methods creating a simple yet effective solution aimed at seabed mapping. 
Sec. \ref{sec:experimental_results} shows the experimental evaluation of the method with data from both real and simulated surveys.

\section{RELATED WORK} 
\label{sec:related_work}
SLAM techniques for autonomous platforms have been applied to map large areas with great success within ground robotics.  
\cite{bosse2004simultaneous}, \cite{jeong2012pushing} are two examples of how algorithmic and hardware-related advancements have enabled autonomous mapping of large regions in real world environments. 
However, the specific challenges posed by the underwater domain have made this branch of mobile robots lag behind its ground counterpart.
Limited sensing and communication capabilities and changing environment dynamics remain challenges for successful underwater deployments of SLAM systems.
The typical AUV navigation sensors suite consist of an inertial measurement unit (IMU) and a Doppler velocity log (DVL) used to compute a dead reckoning (DR) estimate of the vehicle's pose \cite{leonard2016autonomous}.
Exteroceptive sensing underwater mainly relies on sonar technology or cameras to build a consistent model of the seabed while bounding the rapidly growing drift error in the positioning.

Attempts have been made to adapt the SLAM solutions deployed on ground and indoor robotics to AUV technology.
One of the first steps towards underwater SLAM for AUVs can be traced back to \cite{williams2000autonomous}.
Here an extended Kalman filter (EKF) was used to build a feature-based map of the surroundings through the point features extracted from sonar scans.
In \cite{eustice2005visually} the estimation problem is reformulated in terms of extended information filters (IF) to reduce the computational burden of the EKF updates.
The approach is then tested in the context of a vision-based shipwreck survey.
Both works are examples of SLAM systems which depend on the ability to detect and disambiguate features from the environment.
In an underwater surveying context, this is not always a possibility due to the scarcity of these features, low saliency or the limited sensing capabilities.
To overcome these issues, \cite{roman2005improved} introduces a featureless approach based on MBES submaps.
Here, raw MBES scans are collected into patches or seabed submaps, which are associated to a single vehicle pose and treated as a measurement within an EKF.
In this case, data association is carried out by registering overlapping submaps with a variant of the iterative closest point (ICP) algorithm.
In \cite{bore2018sparse}, Bore et al. use these submaps to train sparse Gaussian processes (GP) which are used to reconstruct the final map.
The map refinement is then performed by finding the vehicle poses that maximize the probabilities of overlapping MBES returns to be originated from the corresponding GPs.
The submapping approach has yielded promising results on mid-scale surveys over areas with a rich topology. 
For this reason it has been applied within the framework presented in this paper, and it will be discussed more in detail in \ref{sub:map_consistency_maximization}.
In \cite{fairfield2008active}, Fairfield uses a particle filter to perform localization relative to a grid-based representation of the sea bottom within an active SLAM framework.
Barkby et al. report in \cite{barkby2012bathymetric} on a SLAM system based on a Rao-Blackwellized particle filter (RBPF) in which the weighting of the particles is based on the self-consistency of their 2.5D maps, modeled by GPs.
The measure of the seabed geometric consistency they use, defined in \cite{roman2006consistency}, has also been applied in this work, and will be presented more in detail in section \ref{sub:map_consistency_maximization}.

Factor graphs have been used in underwater SLAM within the context of autonomous mapping of submerged structures \cite{palomeras2018autonomous}, \cite{ho2018virtual}, vehicle localization through sparse bathymetric mapping with a DVL \cite{bichucher2015bathymetric} or flooded inland mine mapping as in \cite{bleier2017signed}. 
But notably its most consistent application in recent years has been in the field of autonomous ship hull inspection.
This kind of operation aims to build a representation of the survey target, the hull of a vessel, while being able to estimate the trajectory of the AUV so as to ensure full coverage.
To do so, Hover \cite{hover2012advanced} combined features extracted from both camera feedback and an imaging sonar to create spatial constraints between vehicle poses in a pose-graph framework. 
In \cite{ozog2016long}, similar constraints are derived from the relative alignment of planar patches associated to an AUV pose and pairwise visual constraints from a camera.
These patches are constructed from the returns of a DVL pointing at the hull and allow for a robust registration in multisession SLAM.
At the expense, however, of requiring a prior estimate of the curvature of the hull.
The work in \cite{teixeira2016underwater} achieves similar results in the same context but rely on the registration of volumetric submaps created from carefully processed MBES scans to build the graph.

Probably the most closely related approaches from an algorithmic perspective to the work presented here can be found in \cite{hover2012advanced}-\cite{teixeira2016underwater}.
However, these solutions have been designed and tested in very different environments to the ones targeted in this paper.
A ship hull on a harbor constitutes a much more structured setup in which smaller dynamics can be expected and assumptions on the geometry of the targets can safely be made, as in \cite{ozog2016long}. 
Furthermore, the sonar propagation is simpler to model than at deeper rates.
And finally, the scale of the surveying areas and vehicle trajectories of all the previous works mentioned are several orders of magnitude different than those of the results presented here.

\section{SLAM WITH BATHYMETRIC SUBMAPS}
\label{sec:slam_bathymetry}
\begin{figure}[!h]
	\centering
	\includegraphics[width=0.4\textwidth]{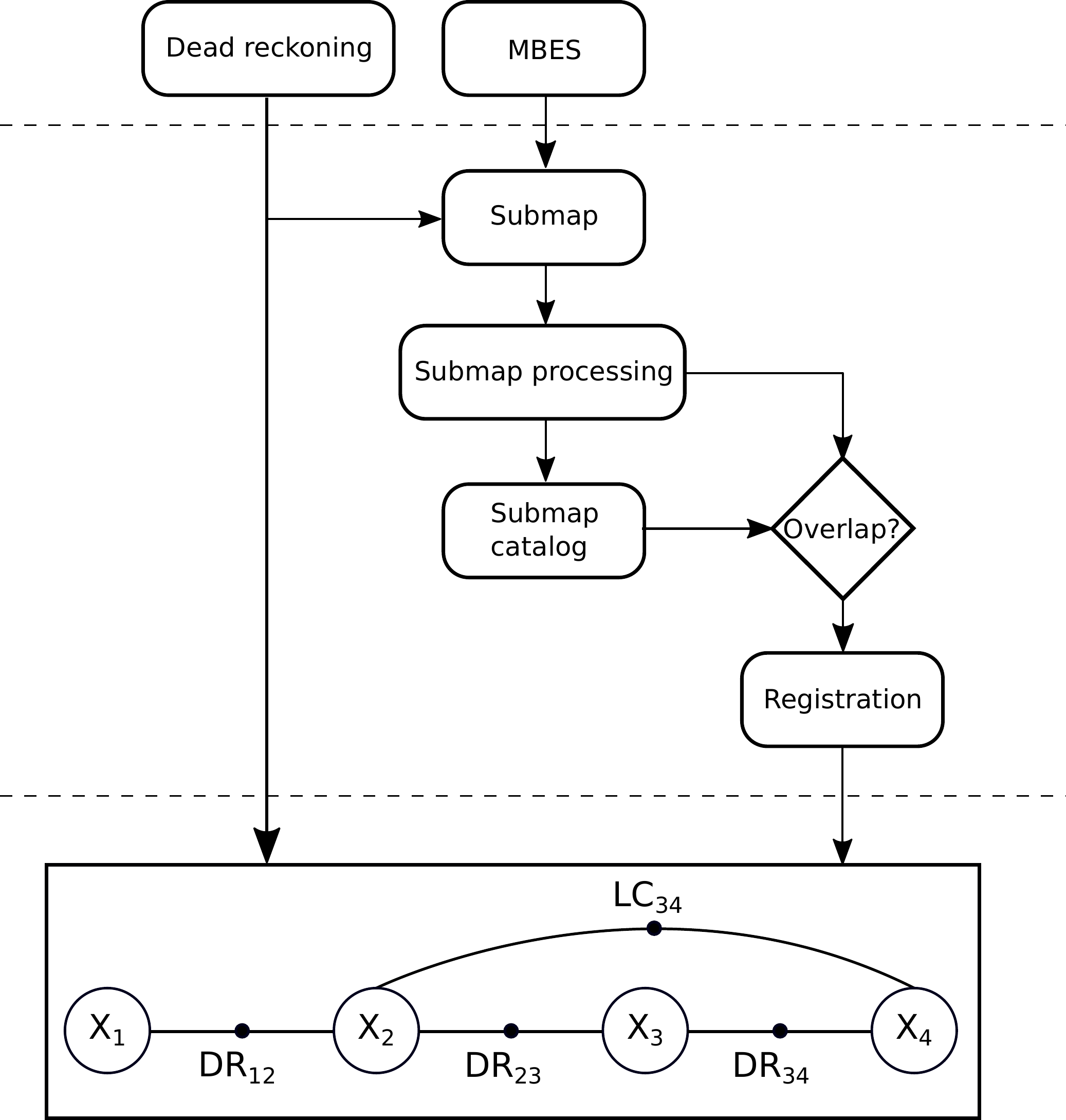}
	\caption{The two levels of our SLAM framework}
	\label{fig:diagram}
\end{figure}

A surveying platform accumulates a growing error in its pose estimate which, together with the resolution of the sensors, will limit the quality of the map being constructed.
On an ROV, this error will be proportional to its distance to the USBL system on the support vessel.
In the case of an AUV, the navigation drift will grow unbounded with the duration of the mission due to the absence of global references.
However, underwater platforms are able to achieve higher accuracy on the bathymetry since they can operate very close to the seabed.
With ship hull-mounted MBES systems, the positioning error is much smaller thanks to the availability of GPS but at the cost of limited resolution.
Thus, a SLAM solution is required to guarantee a level of accuracy on the survey. 
As introduced in \cite{roman2007self}, the geometric consistency of the seabed model and the vehicle trajectory error are related by the sensor measurements.  
This means that maximizing the map consistency will lead to a more accurate trajectory estimate, and vice versa, where the limit is imposed by the accuracy of the sensors.
Thus, we pose the problem of refining the vehicle's trajectory and the map as an optimization problem solved in two steps.
The first step, described in section \ref{ssub:map_local_optimization}, registers locally overlapping MBES swaths over an area and propagates that correction to the estimate of the vehicle pose while collecting those swaths.  
The second step, in \ref{sub:graph_slam}, globally reduces the disparity between the vehicle DR estimate and the corrections from step one.
The diagram of the algorithm described hereafter is shown in Figure \ref{fig:diagram}. 

\subsection{Maximization of the map consistency} 
\label{sub:map_consistency_maximization}
The first step consists in minimizing the consistency error in the seabed map, as the vehicle collects MBES returns. 
To do so, the map inconsistency is computed in all the overlapping areas of the map and then the error is recursively optimized on each of them.

\subsubsection{Error metric} 
\label{ssub:error_metric}
Given the absence of ground truth in underwater mapping, an error metric needs to be defined to measure the consistency of a bathymetric map. 
We selected the geometric disparity defined in \cite{roman2006consistency} given its broad use in similar applications \cite{roman2007self}, \cite{barkby2012bathymetric}.
Intuitively, this measure quantifies the thickness of the point cloud resulting from merging overlapping MBES returns from different swaths over the same area.

\subsubsection{Composition of MBES pings} 
\label{ssub:measurement_composition}
\begin{figure}[!h]
	\centering
	\includegraphics[width=0.45\textwidth]{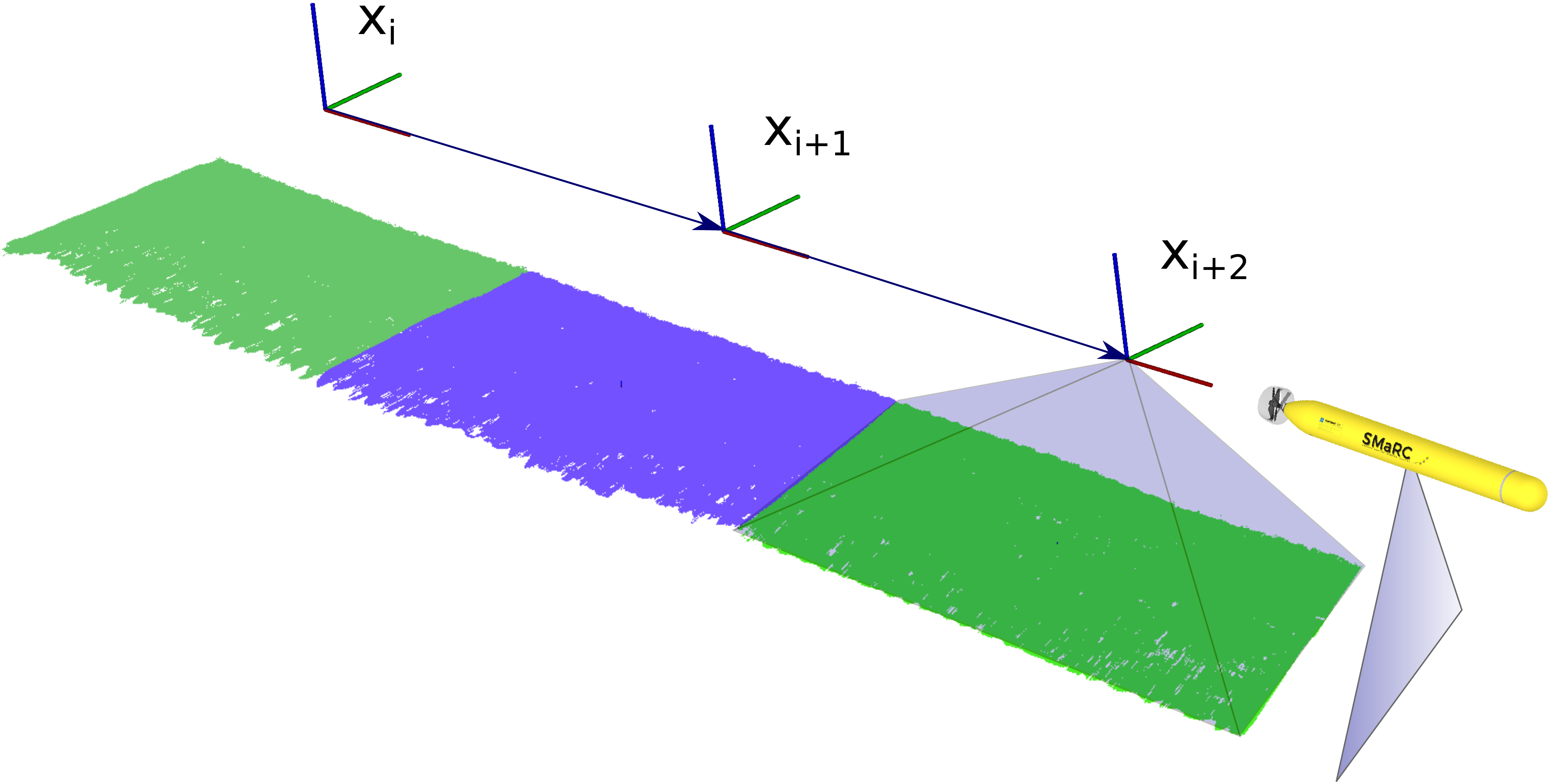}
	\caption{Composition of MBES pings into submaps.}
	\label{fig:sam}
\end{figure}

Accumulating consecutive sonar pings into a submap $S_{i}$ as in
\cite{roman2005improved} allows treating the submaps as single
measurements.  These measurements are then associated to a submap
frame, $x_{i} = [x,y,z,\psi,\phi ,\theta]$ initially given by the dead-reckoning estimate of the vehicle 6D pose while acquiring the ping in the middle of the submap, as depicted in Fig.~\ref{fig:sam}.  The reason to do this is twofold: i) parts of the map can be treated separately according to the magnitude of the
consistency error in the area and ii) measurement
compositions reduce the size of the pose graph being built, described in the next section.  However, for this approach to work the
consistency within the submaps must be preserved.  Hence, submap
formation is triggered according to two criteria:  i) the submap already
contains enough geometric features to contribute meaningful
information to the following map optimization process, described in
\ref{ssub:map_local_optimization} or ii) the submap length has reached a
maximum manually set above which the vehicle trajectory can no longer be assumed to
be drift-free.  Once a submap has been created, a 3D voxel grid is
applied to down-sample it, preserving the geometric information while
reducing the size of the submap.  This is followed by an outlier
removal filter meant to eliminate spurious MBES returns.  These steps
are of key importance when working with large areas of seafloor.


\subsubsection{Local optimization of the map} 
\label{ssub:map_local_optimization}
Once some overlap has been detected among submaps, the latest submap is re-aligned relative to all the others, which are assumed rigidly linked according to the current solution.
A version of the generalized iterative closest point (GICP) from \cite{segal2009generalized} restricted to x,y and yaw is applied to find the relative transformation $T^{GICP} \in \mathbb{R}^3$ between the submap frames that minimizes the plane-to-plane distance between the point clouds.
GICP is more suitable for the registration of large, mostly flat submaps \cite{torroba2018comparison} than the variants of ICP typically used on other more structured underwater environments (\cite{palomeras2018autonomous}, \cite{roman2007self}, \cite{teixeira2016underwater}).
This is because GICP models locally the surface from both point clouds during the matching step, easing the registration of large planar surfaces with minimal overlap and scarce features.    
The output of the re-alignment of submaps $i$ and $j$ is a relative rigid transformation with Gaussian model $T(x_{i}, x_{j}) \sim (T^{GICP}_{i,j}, O_{i})$ that, applied to the source submap, will minimize the distance between corresponding matches among submaps, locally reducing the consistency error of the global map.
However, given that GICP locally optimizes over a non-convex function, its convergence to a global minima is not always guaranteed.
Most algorithms that make use of ICP-based registration methods handle this explicitly by manually setting thresholds on the values of the results or preprocessing the submaps.
Since these thresholds are usually case-dependent, they are a step away from a general algorithm.  
In our case, the second step of the algorithm corrects erroneous outputs of the registration.

\begin{figure*}[!t]
	\centering
	\vspace*{0.05in}
	\includegraphics[trim=0 4 0 5,clip, width=0.95\textwidth]{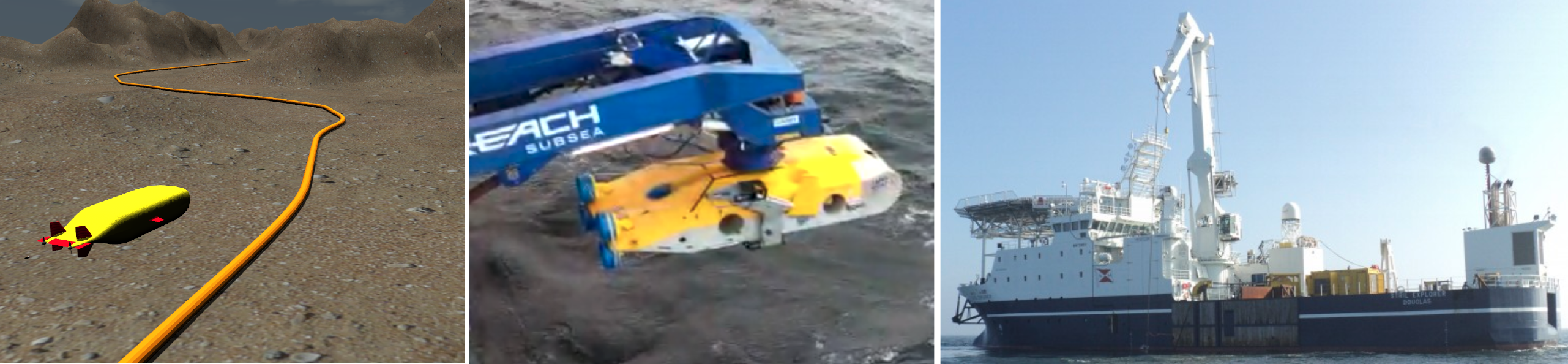}
	\caption{From left to right: SMARC simulation environment, ROV Surveyor and Stril Explorer vessel.}
	\label{fig:platforms}
\end{figure*}

\subsection{Minimization of the vehicle trajectory error} 
\label{sub:graph_slam}
On the second step of the SLAM algorithm, the estimate of the vehicle trajectory is optimized performing maximum a posterior (MAP) inference on a pose graph built from the nonlinear measurements collected during the first step.
By doing so the algorithm also maximizes the self-consistency of the bathymetric map. 
The use of graphical models in the robotics literature in order to solve SLAM as an inference problem can be traced back to \cite{lu1997globally}.
Here, we use a pose graph in which the set of vertices are the $[x,y,\theta]$ components of the submap frames $X = \{x_{0}, ...., x_{N}\}$ and where edges are virtual measurements $z_{i,j}$ relating vertices $x_{i}$ and $x_{j}$, with associated information matrix $\Omega_{i,j}$.
Defining the prediction of a virtual measurement between $x_{i}$ and $x_{j}$ as their expected relative transform such that $\hat{z}_{i,j} = T(x_{i}, x_{j})$, we can define a term to account for the errors present in the system as
\begin{equation}
\label{eq:error_term}
	e_{i,j}(x_{i}, x_{j}) = z_{i,j} - \hat{z}_{i,j}(x_{i}, x_{j})
\end{equation}
In our problem, two sources of virtual measurements are considered when building the graph:

\subsubsection{Dead-reckoning constraints} 
\label{ssub:dead_reckoning_constraints}
A DR estimate of the vehicle positioning with Gaussian model $z_{i,i+1} \sim (f(x_{i}, x_{i+1}), \Sigma_{i})$ is used to compute the virtual measurements between contiguous vertices, resulting in an odometry chain of vehicle poses.  We will refer to this as a DR constraint.
Hence,  eq.~(\ref{eq:error_term}) when $j= i + 1$ becomes
\begin{equation}
\label{eq:dr_term}
	e_{i,j}(x_{i}, x_{j}) = f(x_{i}, x_{i+1}) -  T(x_{i}, x_{j})\; \quad \textrm{with} \quad \Omega_{i,j} = \Sigma^{-1}_{i}
\end{equation}
\subsubsection{Loop closure constraints} 
\label{ssub:loop_closure_constraints}
The GICP registration between overlapping submaps $i$ and $j$ is used to add a loop closure (LC) constraint between the corresponding submap frames, such that the eq.~(\ref{eq:error_term}) for $j \neq i + 1$ becomes
\begin{equation}
\label{eq:lc_term}
	e_{i,j}(x_{i}, x_{j}) = T^{GICP}_{i,j} -  T(x_{i}, x_{j})\; \quad \textrm{with} \quad \Omega_{i,j} = O^{-1}_{i}.
\end{equation}
It shall be mentioned here that the rotational components in equations \ref{eq:dr_term} and \ref{eq:lc_term} are parameterized in the solver as quaternions to avoid singularities.

Given that all the virtual measurements follow Gaussian distributions, each edge of the graph becomes a negative log-likelihood factor $\phi_{i,j}$ of the form 
\begin{equation}
\label{eq:edge}
	\phi_{i,j} = \frac{e_{i,j}^{T} \Omega_{i,j} e_{i,j}}{2}
\end{equation}
Hence, a MAP estimate of the vehicle poses $x_{i,j}$ can be computed through the standard formulation 
\begin{equation}
\label{eq:map_estimate}
	X^{MAP} = \argmin_{x} \sum_{k=1}^{K}\phi_{i,j}
\end{equation}
Where K is the total number of constraints in the graph, including both DR and LC.
An pose graph resulting from this process is shown in Fig.~(\ref{fig:ripples_map}), with LC and DR constraints represented with blue and beige arrows respectively.

The general optimization solver Ceres \cite{ceres-solver} is used to solve eq.~(\ref{eq:map_estimate}), and the resulting vehicle poses are used to update their corresponding submap measurements, which are then merged into the final global map.




\section{EXPERIMENTAL RESULTS}
\label{sec:experimental_results}
In this section, the performance of the proposed method has been tested on both simulated and real data.
Each dataset presented contained an initial consistency error on the final bathymetric map resulting from navigation drift and spurious sonar pings.
For each case, the algorithm introduced has been instantiated with the parameters corresponding to the vehicles and sensors used in the mission and its efforts to minimize this initial error, compiled in table \ref{table:results}.

\begin{table}[h]
\caption{Datasets characteristics}
\label{table:datasets}
\begin{center}
\begin{tabular}{|c|c|c|c|c|}
\hline
Dataset & Simulation & Pipeline & Loop  & Ripples \\
\hline
Area (km$^{2}$) & 0.12 & 0.0585 &  0.21 & 1.04 \\
\hline
Swaths & 2 &  7 & 23 & 10 \\
\hline
Time (h) & - & 0.8  & 1.6 &  4.3\\
\hline
Sonar pings & 319,858 & 40,072,188 & 39,237,119 & 32,366,912 \\
\hline
\end{tabular}
\end{center}
\end{table}

\subsection{Simulation environment} 
\label{sub:simulation_test}
Our SLAM algorithm has been developed and initially tested within the SMARC underwater simulator, an open source simulation environment for AUVs based on Gazebo and the Robot Operative System (ROS).
An AUV with a MBES surveys a target area, shown in Figure \ref{fig:platforms}, on two swaths with approximately 30\% overlap.
The robot trajectories are corrupted with accumulative white Gaussian noise with mean zero, added between submap frames so as to preserve the positioning within submaps error free. 
The consistency error in the bathymetry map has been computed before and after the application of the SLAM algorithm.
The results for one run are shown in table \ref{table:results} for reference.
Both the simulator and the framework code are available online.


\begin{figure*}[!ht]
\center
\begin{subfigure}[t]{.47\linewidth}
	\begin{tikzpicture}[scale=0.7]
		\node(map) {\includegraphics[width=0.95\linewidth]{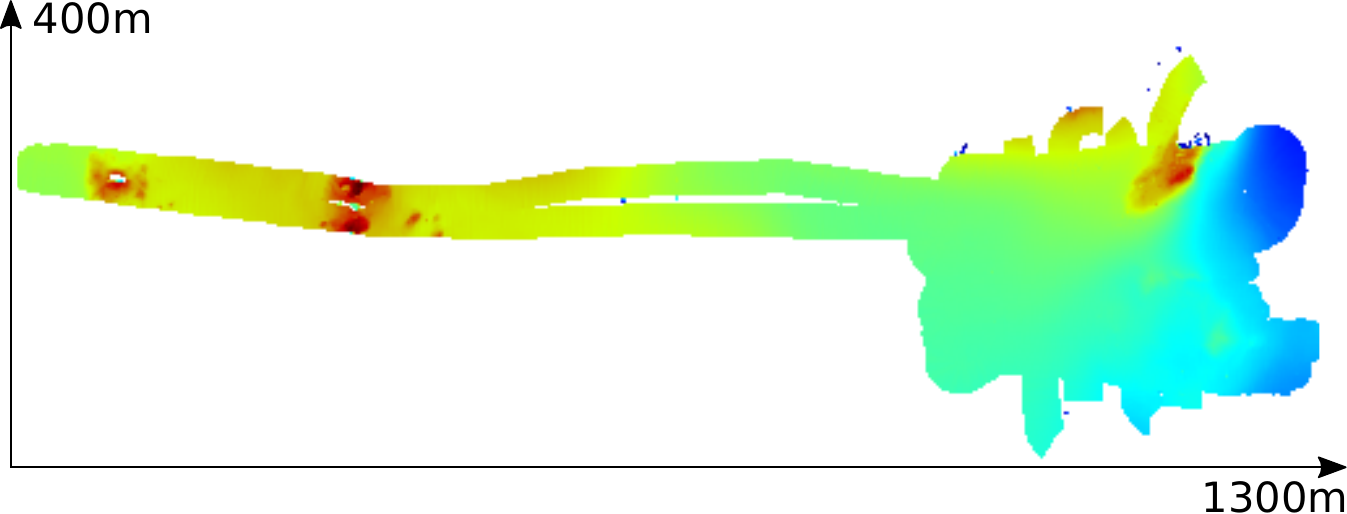}};
	\end{tikzpicture}
	\caption{Initial depth map.}
\end{subfigure}
\begin{subfigure}[t]{.47\linewidth}
    \begin{tikzpicture}[scale=0.7]
		\node(map) {\includegraphics[width=0.95\linewidth]{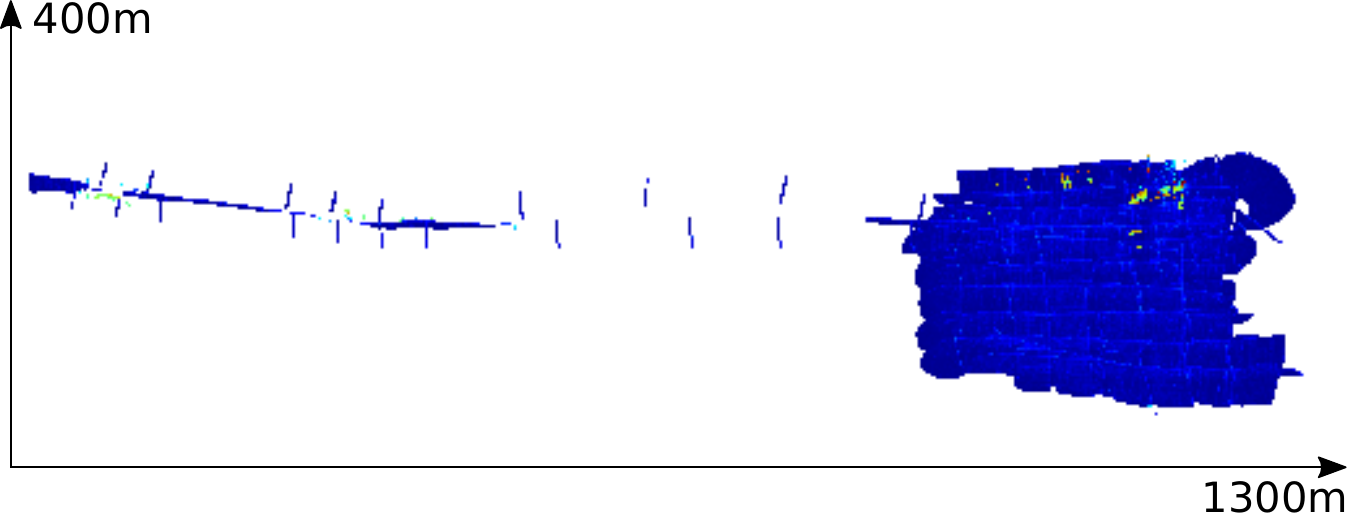}};
	\end{tikzpicture}
    \caption{Initial map of the consistency error on overlapping areas.}
\end{subfigure}
\begin{subfigure}[t]{.47\linewidth}
	\begin{tikzpicture}[scale=0.7]
		\node(map) {\includegraphics[width=0.95\linewidth]{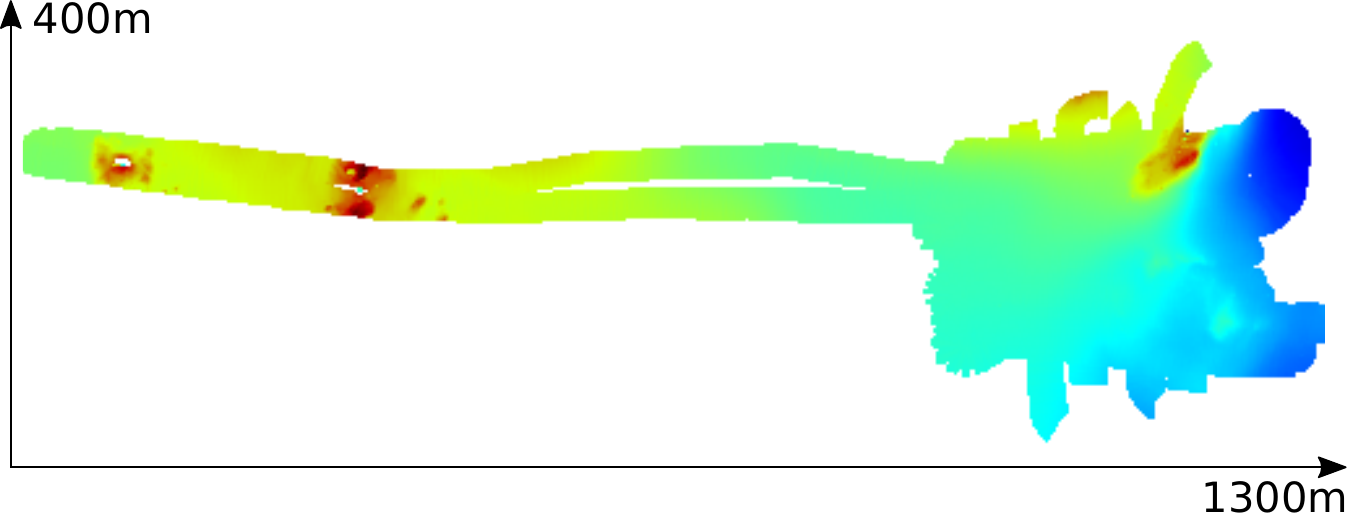}};
	\end{tikzpicture}
    \caption{Depth map after applying our SLAM solution}
\end{subfigure}
\begin{subfigure}[t]{.47\linewidth}
    \begin{tikzpicture}[scale=0.7]
		\node(map) {\includegraphics[width=0.95\linewidth]{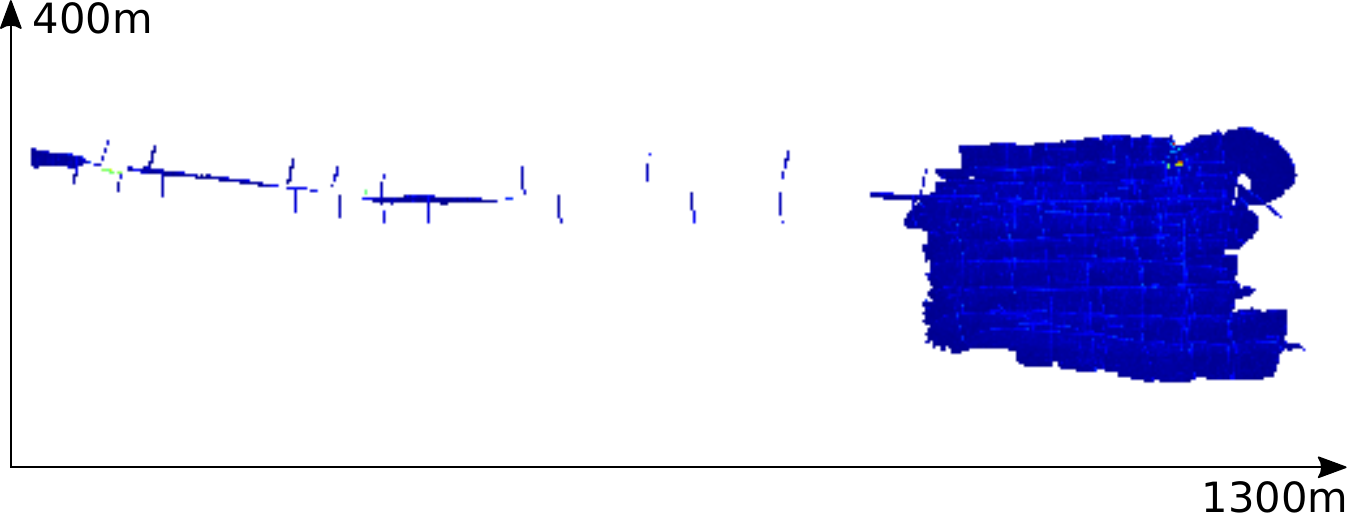}};
	\end{tikzpicture}
    \caption{Consistency error removed, most notably on the top area.}
\end{subfigure}
\caption{Bathymetric and consistency error maps of the Loop dataset before and after applying our SLAM framework.}
\label{fig:loop_dataset}
\end{figure*}

\subsection{Real datasets} 
\label{sub:datasets}
A common characteristic in real industrial bathymetric surveys is that generally they do not present large consistency errors due to the high standards of the equipment used.
However, misalignments in the range of 1-2 meters can be critical when, for instance, assessing the state of underwater infrastructure.
The datasets shown present various degrees of complexity in terms of accumulated navigation error, size of the area covered, resolution and noise of the MBES, percentage of overlap among swaths and saliency of the terrain.
They all contain vehicle positioning, raw MBES returns and sound velocity profile (SVP) data, and they have been collected at three different geographical locations by different platforms.
Their characteristics are presented in \ref{table:datasets}.

\subsubsection{Loop} 
\label{ssub:loop_dataset}
This mission was launched to detect several objects laying on the seabed at a depth rate of 15 meters on average.
Given the small size of the targets, high accuracy on the bathymetric map was a requirement.
Therefore, the vessel with a hull-mounted MBES followed a lawnmower pattern with major overlap between swaths and a big loop towards the end of the trajectory.    
The initial bathymetry map can be seen in Fig.~\ref{fig:loop_dataset} a).
In terms of the pose graph construction, a lawn mowing pattern results in a high concentration of loop closure constraints with major overlap.
Thus, the resulting graph is able to correct erroneous GICP registrations that may occur on flat areas, relaxing the need to fine tune the GICP parameters, as pointed out on \ref{ssub:map_local_optimization}.
The topology of the area is mostly flat with major features mapped on the top and the loop regions, which is where most of the initial consistency error is detected.
The map of the initial consistency error, computed on the areas with overlap between submaps, is shown in the map in Fig.~\ref{fig:loop_dataset} b).
Fig.~\ref{fig:loop_dataset} d) depicts the consistency error map after applying our SLAM algorithm.
It has successfully eliminated the geometric error on the areas with a sharper relief, resulting on a different topology on the final depth map, shown in the map in Fig.~\ref{fig:loop_dataset}. 

\subsubsection{Pipeline} 
\label{ssub:pipeline}
\begin{figure}[!h]
	\centering
	\includegraphics[width=0.45\textwidth]{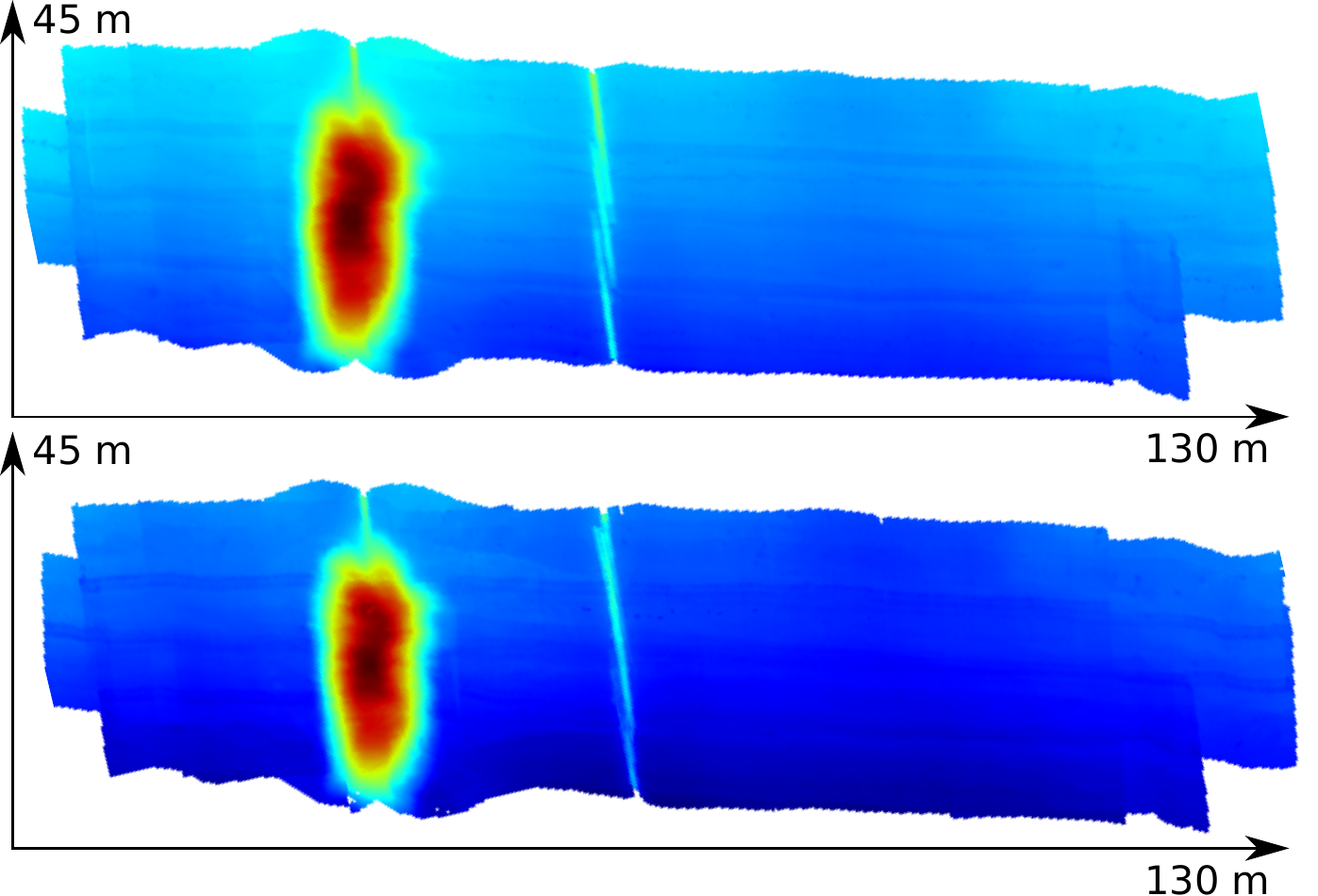}
	\caption{Pipeline dataset with initial misalignment (top) and correction with the SLAM framework (bottom)}
	\label{fig:medgaz}
\end{figure}

This dataset consists of an inspection over a section of a pipeline at approximately 2500 meters depth.
It was collected with a Surveyor Interceptor ROV, shown in Fig.~\ref{fig:platforms}.
Industrial inspection is usually carried out acquiring long, parallel swaths over the area of interest with 20-30$\%$ overlap.
The drift in the ROV positioning among tracks can be easily appreciated in the misalignment of the pipeline segments across swaths and the artifacts on the hill on the top figure on Fig.~\ref{fig:medgaz}, which is clearly reduced on the SLAM solution, at the bottom.  
Our algorithm has divided each swath into submaps according to the criterion in \ref{ssub:measurement_composition}.
The correction of the pipeline has been a result of the registration of the submaps containing the hill, which the graph optimization has propagated through the ROV trajectory.


\subsubsection{Ripples} 
\label{ssub:the_ripples_dataset}
This dataset consists of the search of a large target over an area of interest at an average water depth of 30 meters, shown in Fig.~\ref{fig:ripples_map}.
Its main characteristics are the absence of crossing tracks and the minimal overlap between parallel swaths, since the aim of this kind of missions is to cover as much area as possible.
It was collected with a MBES mounted on the hull of the Stril Explorer vessel, shown on the right of Fig.~\ref{fig:platforms}.
Given that the positioning of the vessel is very accurate, most of the consistency error in this dataset is estimated to arise from noise in the sonar returns. 
The raw bathymetry map presents wavy patterns most likely caused by motion on the pitch angle of the vessel.
These patterns are contained within the submaps and therefore our algorithm could not correct them.

\begin{table}[h]
\caption{RMS consistency error}
\label{table:results}
\begin{center}
\begin{tabular}{|c|c|c|c|c|}
\hline
Dataset & Simulation & Pipeline & Loop & Ripples \\
\hline
Initial & 0.82 & 0.17 & 1.19 & 1.02\\
\hline
Optimized with SLAM & 0.71 & 0.15 & 0.85 & 0.57 \\
\hline
\end{tabular}
\end{center}
\end{table}

\section{CONCLUSIONS \& FURTHER WORK}
\label{sec:conclusions}
MBES-based industrial bathymetric surveys for infrastructure layout or inspection require the highest level of accuracy for safety reasons. 
In this paper, we have presented an algorithm to optimize the self-consistency of the maps produced by this kind of surveys in two steps.
Our system has a component that corrects the latest vehicle pose being added to the map in a filter-like phase by maximizing the self consistency of the map and a component that performs a global optimization over the entire trajectory.  It depends on the DR in regions of low terrain information and on the consistency of the map in areas rich on features.

However, in the current system the performance of the loop closure detection relies on the lawn mowing pattern with partially overlapping swaths that industrial surveys typically follow.
To handle more general cases, accounting for the pose uncertainty in the loop closure detection is key for a more robust solution.
Furthermore, in this work the value of $O_{i}$ is obtained from an heuristic based on the normalized covariance of the returns on the submap, as in \cite{vanmiddlesworth2015mapping}.
In future work this will be replaced for a value that better represents the uncertainty of the GICP output.

Finally, both the SLAM framework and the simulation environment have been released and can be found in \footnote{https://github.com/smarc-project}. 
The \textit{Pipeline} dataset has been made publicly available in \footnote{https://strands.pdc.kth.se/public/IROS-2019-Bathymetry/}.


\section{ACKNOWLEDGMENTS}
The authors want to thank MMT and the Stril Explorer crew for their assistance on the collection of the data used on this work.
This  work  was  supported  by  Stiftelsen för Strategisk Forskning  
(SSF)  through  the  Swedish  Maritime Robotics Centre (SMaRC) (IRC15-0046).

\bibliography{root}
\bibliographystyle{unsrt}

\end{document}